\documentclass{article}

\newcommand{\isaccepted}{}
\usepackage[final]{corl_2026}
\makeatletter
\renewcommand{\@conferencelocation}{Austin, Texas, USA}
\makeatother

\usepackage{graphicx}
\usepackage{multirow}

\usepackage{amsmath}
\usepackage{amssymb}
\usepackage{booktabs}
\usepackage{subcaption}
\usepackage{wrapfig}
\usepackage{float}

\title{SuperSuit: An Isomorphic Bimodal Interface for Scalable Mobile Manipulation}

\author{
  Tongqing Chen$^{1,2}$, Hang Wu$^{2,\dagger}$, Jiasen Wang$^2$, Xiaotao Li$^2$\\
  $^1$Tsinghua University, Beijing, China\\
  $^2$TARS Robotics, Shanghai, China\\
  $^\dagger$ Corresponding author.
}

\hypersetup{
  pdftitle={SuperSuit: An Isomorphic Bimodal Interface for Scalable Mobile Manipulation},
  pdfauthor={Tongqing Chen, Hang Wu, Jiasen Wang, Xiaotao Li}
}

\begin{document}
\maketitle

\begin{figure}[H]
  \centering
  \includegraphics[width=0.66\textwidth]{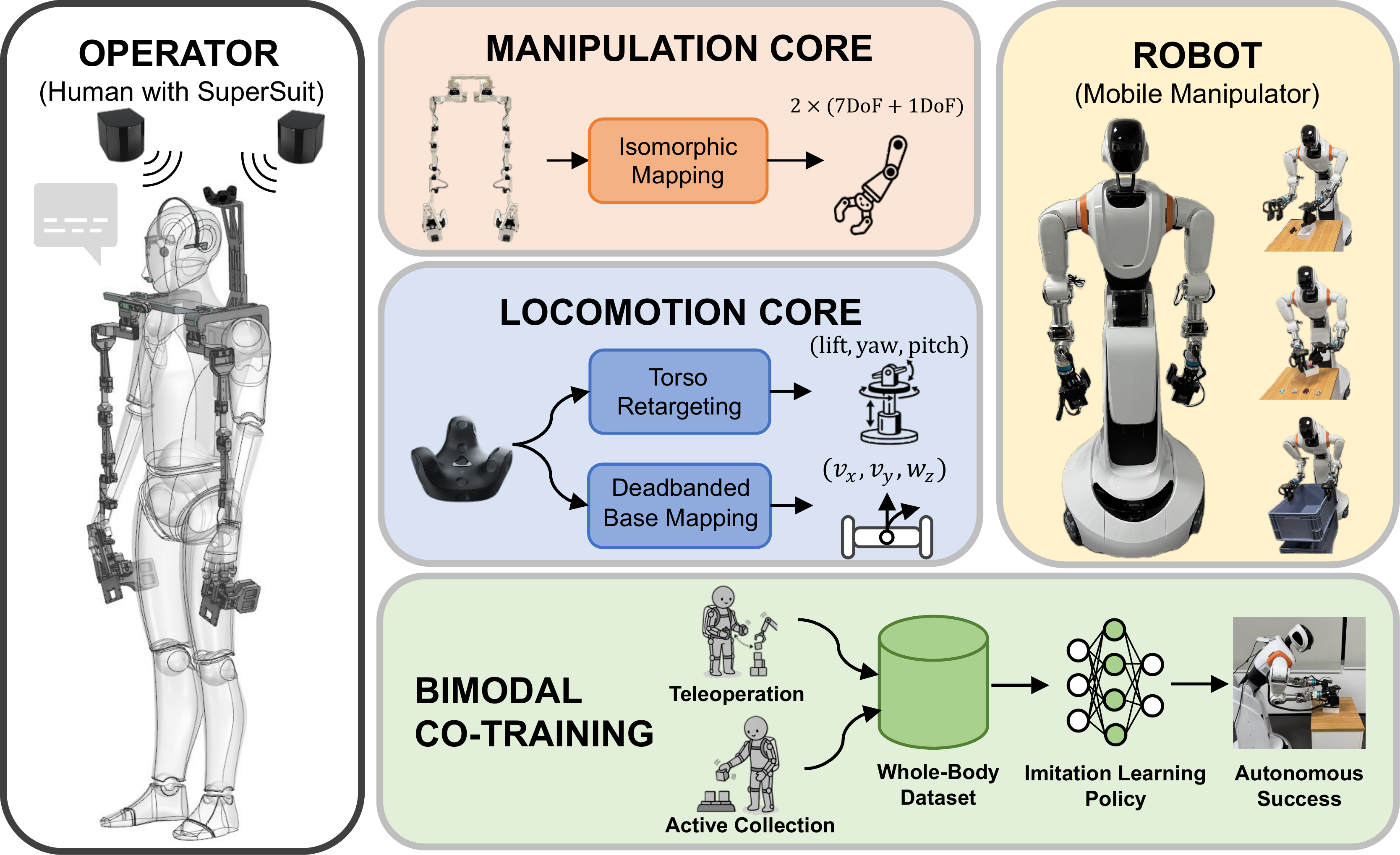}
  \caption{
    \textbf{The SuperSuit Framework.}
    Our untethered wearable interface translates human embodiment to whole-body robot control via isomorphic arm manipulation and deadbanded base locomotion.
    Both teleoperation and active collection produce demonstrations for imitation learning.
  }
  \label{fig:teaser}
\end{figure}

\begin{abstract}
  Long-horizon mobile manipulation needs whole-body demonstrations, and collecting them is the bottleneck. Keeping the robot in the loop ties the dataset to robot uptime, while recording human motion without it instead requires a conversion step that decides the joint-space action the policy learns.
\textbf{SuperSuit} is a single wearable that both teleoperates a wheeled bimanual robot and, with the robot absent, records demonstrations for it.
Its arms are mechanically aligned to the robot's, so recorded joint motion needs no inverse kinematics and no retargeting; one body-worn tracker drives the torso and the planar base through a deadband that keeps posture micro-sway out of the base command.
Both roles emit the same 22-D command over the torso, planar base, two 7-DoF arms, and two grippers, with arm targets expressed as forward-looking increments of the observed joint state.
On three real-robot tasks this lets active demonstrations replace most robot-in-the-loop collection without changing policy quality. With a 10-episode robot-grounded seed held fixed, the remaining 100 episodes can be recorded without the robot, whereas active-only training falls well short.
SuperSuit moves the limit on dataset size from the robot to the operator.

\end{abstract}


\section{Introduction}
Imitation learning is mature for short-horizon manipulation~\cite{act, diffusion, pi_zero, pifast, smolvla, openvla, openvlaoft, 4d-vla}, but long-horizon mobile manipulation needs demonstrations that coordinate base motion, torso articulation, and bimanual action~\cite{openx, robocoin, khazatsky2024droid}, and acquiring them is the bottleneck.
Wheeled platforms offer a large planar workspace~\cite{stretch, mobilealoha}, and two routes dominate whole-body data collection on them: keep the robot in the loop for every demonstration, or record human motion without it and transfer that motion to robot actions afterward.

Keeping the robot in the loop gives genuine robot observations, but it ties the amount of data to how often the robot is free. Every episode consumes robot uptime, resets, and recovery, and the recorded motion must stay inside the platform's stability and safety envelope~\cite{mobilealoha, brs}.
Recording human motion removes that cost and substitutes a conversion step.
End-effector and wrist-tracked capture such as UMI~\cite{umi, fastumi, umi-ft, dexumi} records a task-space proxy instead, so the action the policy learns comes out of a retargeter rather than out of the demonstration.
Isomorphic exoskeletons~\cite{worobot, airexo, airexov2, homie, nuexo, humanoidexo} avoid that retargeter, but none carries a wheeled base, where planar base motion is itself part of the action.
Neither route produces one schema in which the robot-in-the-loop and robot-free episodes of such a platform are jointly trainable (Sec.~\ref{sec:related}).

SuperSuit is built around an acquisition-time action contract. Both collection modes are fixed to emit the same action before any policy is trained.
Each produces a 22-D command over the torso, planar base, two 7-DoF arms, and two grippers.
The arms are mechanically aligned to the robot, so joint correspondence is direct rather than retargeted, and arm targets are forward-looking increments relative to the observed joint state~\cite{srt}.
Base velocity is deadbanded so that small posture shifts are not recorded as locomotion.
Robot-grounded and robot-free episodes therefore enter training under the same action supervision, with no per-mode conversion and no post-hoc trajectory retargeting.

Active collection records more than twice the episodes per measured collection hour of a BRS-style teleoperation baseline across three long-horizon real-robot tasks (Sec.~\ref{subsec:task_setup}, App.~\ref{app:statistics}).
With 10 robot-grounded episodes held fixed, replacing the other 100 with active demonstrations leaves success unchanged within the resolution of our rollouts, relative to 110 teleoperation episodes, whereas active-only training succeeds in only 35\%, 10\%, and 5\% of rollouts.

Our contributions are:
\begin{itemize}
  \item An acquisition-time whole-body action contract for wheeled bimodal collection: robot-in-the-loop and robot-free demonstrations produce the same 22-D schema, with direct arm-joint correspondence, matched command semantics and temporal references, and no modality-specific retargeting.
  \item A whole-body control realization: forward-looking $\Delta q$ arm targets that cancel a shared constant joint-coordinate offset, a locomotion detector that splits the single measured yaw between torso and base, and deadbanded, clipped base velocities in a fixed reference frame.
  \item Empirical validation that bimodal collection scales: active collection more than doubles the throughput of the BRS-style baseline, and a dataset that needs the robot for only a small share of its episodes matches the policy quality of fully teleoperated data, while active-only data falls clearly short.
\end{itemize}

\section{Related Work}
\label{sec:related}
\subsection{Robot-in-the-Loop Teleoperation for Data Collection}
\label{teleop_collection}

Robot-in-the-loop teleoperation collects whole-body data on both humanoid and wheeled platforms.
Humanoid systems pair immersive interfaces with full-body retargeting under balance constraints~\cite{humanplus, expressive, omnih2o, clone, amo, twist, twist2}, so every movement the operator makes is first solved into a joint trajectory that keeps the robot upright.
Wheeled platforms such as Mobile ALOHA~\cite{mobilealoha} and BRS~\cite{brs} command the hardware directly instead, and in BRS the base and torso are driven through a handheld controller while the arms follow the exoskeleton, so whole-body motion is assembled from two interfaces rather than captured as one.
In both cases the ceiling is set by the robot rather than by the operator: how much data a session yields depends on how long the hardware can keep running.

\begin{figure*}[t]
\centering
\includegraphics[width=1.0\textwidth]{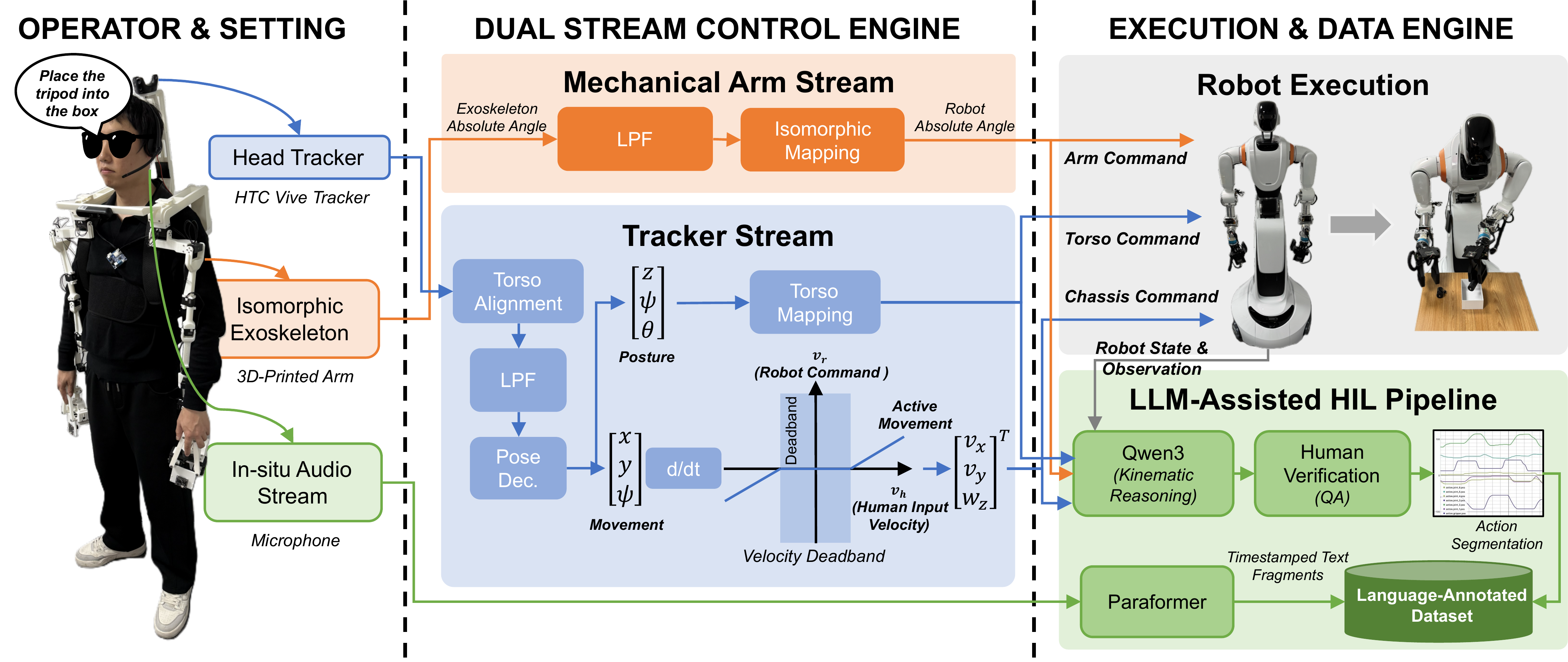}
\caption{
\textbf{System Architecture of SuperSuit.}
The wearable signals are separated into a mechanical-arm stream for isomorphic arm mapping and a tracker stream for torso and base control. The tracker pose is converted to torso targets and fixed-frame planar-base velocities, followed by a velocity-level deadband. The commands drive the robot during teleoperation and are logged in both collection modes for the offline annotation pipeline, where an LLM combines Paraformer transcriptions with kinematic events to propose subtask boundaries that a human annotator verifies and corrects.
}
\label{fig:system_overview}
\end{figure*}

\subsection{Robot-Free and Embodied Demonstration Interfaces}
\label{robot_free_collection}

Robot-free interfaces such as UMI and its extensions~\cite{umi, fastumi, umi-ft, dexumi, rdt2, onetwovla} record end-effector motion in the operator's own workspace and transfer it to a robot, which takes the robot out of collection altogether.
Whole-body extensions for mobile and humanoid platforms~\cite{dobbe, umi-on-leg, portela2whole, humi, egohumanoid} add base motion or a locomotion controller, but each targets one embodiment and needs its own retargeting.
Retargeting decides the joint-space action the policy is trained on, so an error in the mapping is learned as supervision rather than showing up as collection noise.

Isomorphic exoskeletons remove the retargeter by aligning joints mechanically, but the systems built so far cover the arms alone~\cite{worobot, airexo, airexov2} or a humanoid body~\cite{homie}, and leave out a wheeled base.
NuExo combines humanoid teleoperation with robot-independent upper-limb capture~\cite{nuexo}, though its action stops at the upper limb and it reports neither a shared whole-body representation nor mixed-source imitation learning; HumanoidExo does report both robot-free whole-body learning and mixed teleoperation/exoskeleton training, but on a humanoid~\cite{humanoidexo}.
Neither covers a wheeled bimanual platform.
SuperSuit records the same torso, base, arms, and grippers (22-D) in both modes, with arm increments invariant to matched constant joint-coordinate offsets, so one policy consumes either source without a per-mode action space or a separate retargeter.

SuperSuit also adopts two representation choices that are already established elsewhere: relative arm actions, which keep the targets robust to imperfect kinematic measurement~\cite{srt}, and language-assisted annotation, which proposes subtask boundaries from narration that a person then verifies~\cite{nils}.

\section{Methodology}
\subsection{System Overview}

\begin{figure}[t]
\centering
\includegraphics[width=0.8\textwidth]{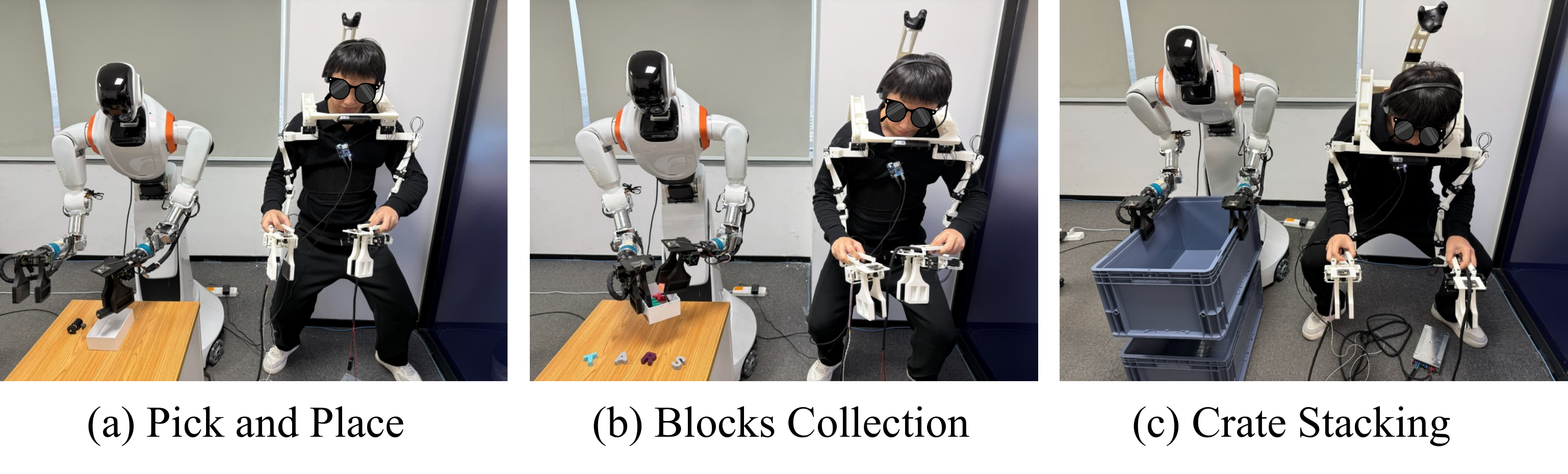}
\caption{\textbf{Remote Teleoperation.} The operator drives the robot in real time with SuperSuit in (a) Pick-and-Place, (b) Blocks Collection, and (c) Crate Stacking.}
\label{fig:teleop_mode}
\end{figure}

SuperSuit (Fig.~\ref{fig:system_overview}) is a bimodal data-collection and teleoperation system for wheeled mobile manipulators.
The operator wears a 3D-printed exoskeleton whose arm joints correspond to those of the target robot.
An HTC Vive Tracker, localized by HTC Base Station 2.0 lighthouses, provides pose estimates for torso and planar-base commands; it sits above the operator's head on a bracket fixed to the back of the suit, and a headset microphone records in-situ narration.

\begin{figure}[t]
\centering
\includegraphics[width=0.8\textwidth]{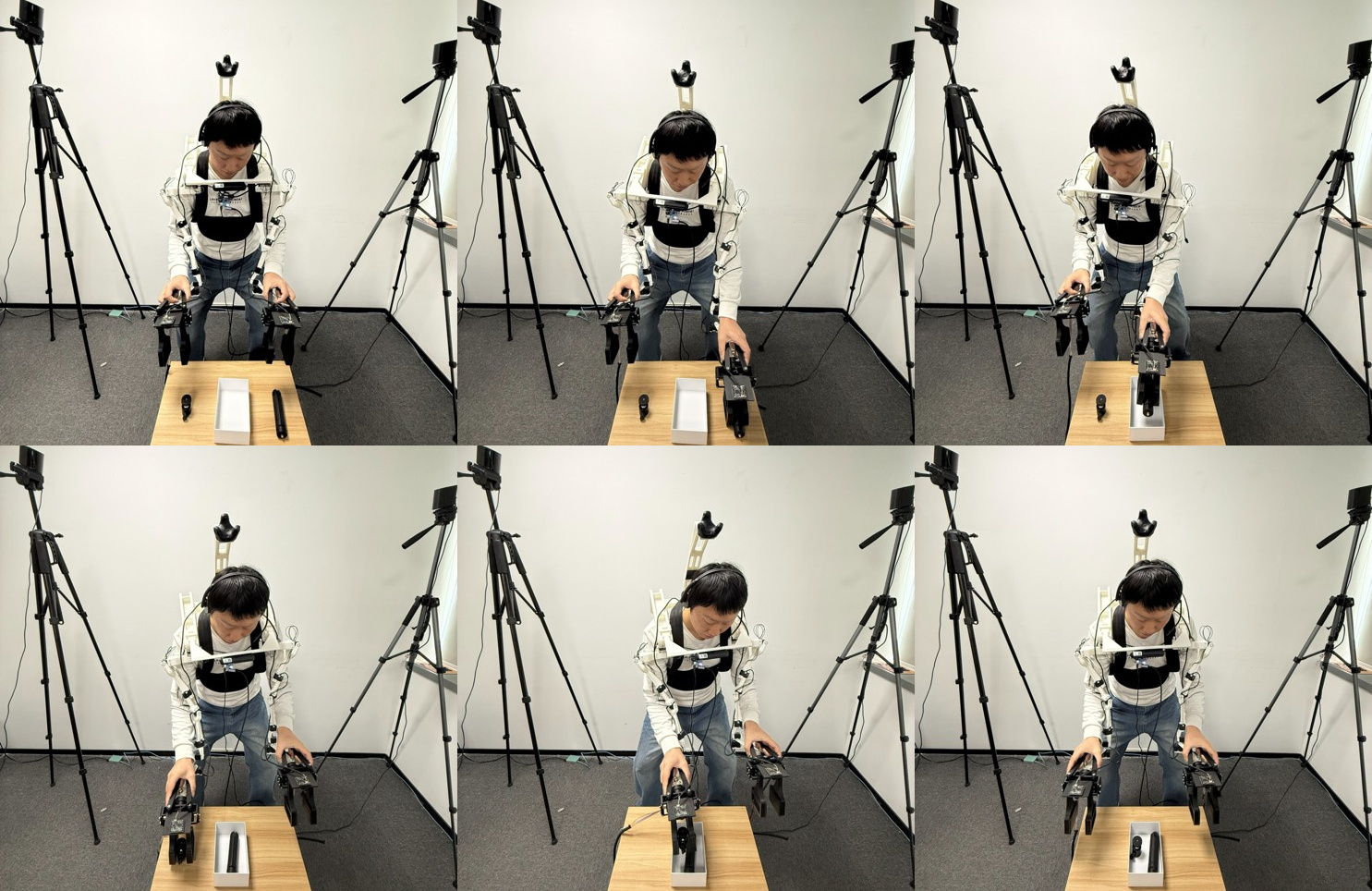}
\caption{\textbf{Active Demonstration.} Robot-free collection under the same whole-body action schema.}
\label{fig:active_mode}
\end{figure}

The system runs in two collection modes (Fig.~\ref{fig:teleop_mode}, Fig.~\ref{fig:active_mode}).
In \emph{Remote Teleoperation} the robot executes the mapped commands live; in \emph{Active Demonstration} the same commands are logged while the operator performs the task directly, so the robot is not needed during recording.
Both modes emit the same torso, base, arm, and gripper variables under one action schema.
Narration is recorded in both; timestamped subtask labels are generated offline and verified by a human annotator, and the policies in this study use them only when explicitly identified as $\pi_{0.5}^{+}$.

\subsection{Human-to-Torso-and-Base Kinematic Retargeting}
\label{subsec:locomotion}

The spatial tracker is carried on a bracket above the operator's head, attached to the back of the suit: it stays clear of the arms during bimanual manipulation and leaves head movement unconstrained.
However, the robot reference is defined at the articulated torso (specifically the torso pitch link) and the mobile base.
We therefore perform an offline rigid calibration to align the tracker frame with the robot torso reference frame.
The resulting mapping is fixed once per setup and identical in both collection modes, and it covers the torso and planar base only; arm and gripper commands come straight from the exoskeleton (Sec.~\ref{subsec:isomorphism}).

We use homogeneous column vectors and write ${}^{A}\!\mathbf{T}_{B}$ for the transform mapping coordinates from frame $B$ to frame $A$. Let $\mathbf{T}_h(t)={}^{W}\!\mathbf{T}_{H}(t)$ be the tracker pose in the world frame and let $\mathbf{T}_{cal}={}^{H}\!\mathbf{T}_{T}$ be the fixed mounting transform from the tracker-derived torso reference frame to the tracker frame. The torso reference pose is therefore ${}^{W}\!\mathbf{T}_{T}(t)=\mathbf{T}_h(t)\mathbf{T}_{cal}$. At initialization time $t_0$, we record this pose and express subsequent commands in the fixed frame $\mathcal{F}_0$ coincident with the torso reference at $t_0$:
\begin{equation}
 {}^{0}\!\mathbf{T}_{T}(t)=\left[{}^{W}\!\mathbf{T}_{T}(t_0)\right]^{-1}{}^{W}\!\mathbf{T}_{T}(t).
\end{equation}
We decompose this relative pose into translation and orientation components $(x(t),y(t),z(t),\allowbreak\psi(t),\theta(t),\phi(t))$, where position is expressed in $\mathcal{F}_0$ and $(\psi,\theta,\phi)$ are relative yaw, pitch, and roll. The torso reference is a tracker-derived proxy; its calibration and the suit-to-torso rigidity assumption are specified in App.~\ref{app:tracker_calibration}.

The torso lift and pitch joints use the vertical displacement and pitch components $(z,\theta)$. The single measured yaw $\psi$ is shared between the torso-yaw joint and the base by a locomotion detector $m(t)\in\{0,1\}$ evaluated before command deadbanding. The detector uses filtered planar motion, including yaw motion, so that both translation and in-place turns can activate locomotion. The robot heading is the commanded base heading $\psi_b(t)$, integrated from the base yaw command $\omega_z$ in Eq.~\eqref{eq:deadband_mapping}, plus the torso yaw. The allocation drives this sum toward $\psi$ through the residual $e_\psi(t)=\mathrm{wrap}(\psi(t)-\psi_b(t))$: during stationary manipulation it becomes the torso-yaw target directly, and during locomotion it becomes a proportional heading-error feedback on the base, with gain $1/\Delta t$ in angular rate:
\begin{equation}
\begin{bmatrix}q_{\mathrm{torso,yaw}}^{cmd}(t)\\ \omega_h(t)\end{bmatrix}
=
\begin{cases}
[e_\psi(t),\,0]^T, & m(t)=0,\\
[0,\,e_\psi(t)/\Delta t]^T, & m(t)=1.
\end{cases}
\label{eq:locomotion_gate}
\end{equation}

When locomotion starts, the torso returns to zero and its yaw is transferred to the base rather than lost. Since $\psi_b$ integrates the base yaw command, once the base has caught up, $\psi_b(t)=\psi(t-\Delta t)$ and the feedback term reduces to the finite-difference rate of $\psi$. Base-rate clipping delays this transfer when the residual is large.
The planar translation is differentiated in $\mathcal{F}_0$:
\begin{equation}
v_{h,x}(t) = [x(t)-x(t-\Delta t)]/\Delta t,\qquad
v_{h,y}(t) = [y(t)-y(t-\Delta t)]/\Delta t.
\label{eq:planar_diff}
\end{equation}
The resulting velocities are rotated from the fixed frame $\mathcal{F}_0$ into the torso reference $\mathcal{F}_T$ before being issued as the base translation command. Because $\mathcal{F}_T$ shares its $x$ and $y$ axes with the base frame $\mathcal{F}_B$, no further rotation is needed between the two, and $v_{h,x},v_{h,y}$ in Eq.~\eqref{eq:deadband_mapping} denote the components after this conversion.

Despite temporal smoothing, differentiated human motion still contains small involuntary micro-sway induced by natural posture stabilization.
If directly integrated into base commands, these micro-motions lead to undesirable base jitter during contact-rich manipulation.

We therefore apply a fixed-threshold deadband to the differentiated velocities $v_{h,x},v_{h,y}$ in Eq.~\eqref{eq:planar_diff} and the base yaw rate $\omega_h$ in Eq.~\eqref{eq:locomotion_gate}.
The deadband operator $\mathcal{D}(v,\epsilon)$ zeroes a velocity component with $|v|\leq\epsilon$ and passes larger values through unattenuated. The threshold is hard, so the command steps by at most $\epsilon$ when a component crosses it; the chassis velocity controller tracks its target with finite bandwidth, which keeps the step from reaching the wheels as a jump.

The base command applies the deadband, then componentwise clipping (App.~\ref{app:deadband}):
\begin{equation}
\begin{aligned}
\mathbf{u}^{raw}(t)&=
\begin{bmatrix}
K_x \mathcal{D}(v_{h,x}(t), \epsilon_x)\\
K_y \mathcal{D}(v_{h,y}(t), \epsilon_y)\\
K_\omega \mathcal{D}(\omega_h(t), \epsilon_\omega)
\end{bmatrix},\\
\mathbf{u}^{base}(t)&=\operatorname{clip}(\mathbf{u}^{raw}(t),\mathbf{u}_{max}),
\qquad \mathbf{u}^{base}=[v_x,v_y,\omega_z]^T.
\end{aligned}
\label{eq:deadband_mapping}
\end{equation}

\subsection{Joint Correspondence and Relative Arm Targets}
\label{subsec:isomorphism}

\begin{figure}[t]
\centering
\begin{minipage}[t]{0.335\textwidth}
\vspace{0pt}
\centering
\includegraphics[width=\linewidth]{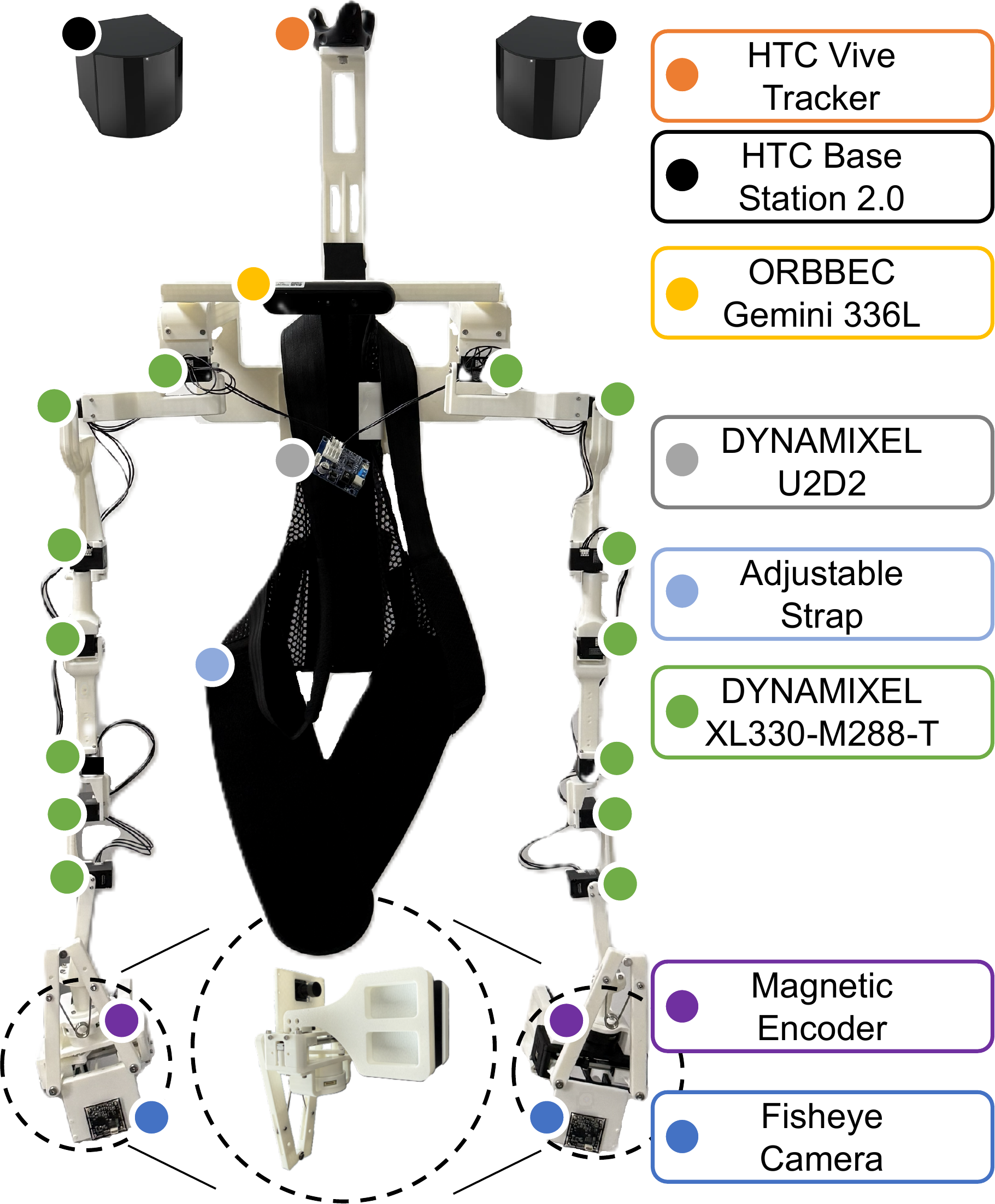}
\caption{\textbf{Kinematic Alignment of the SuperSuit.} The exoskeleton's mechanical axes mirror the operator's anatomical degrees of freedom.}
\label{fig:cad_design}
\end{minipage}\hfill\nextfloat
\begin{minipage}[t]{0.625\textwidth}
\vspace{0pt}
\centering
\begin{subfigure}{\linewidth}
    \centering
    \includegraphics[width=\linewidth]{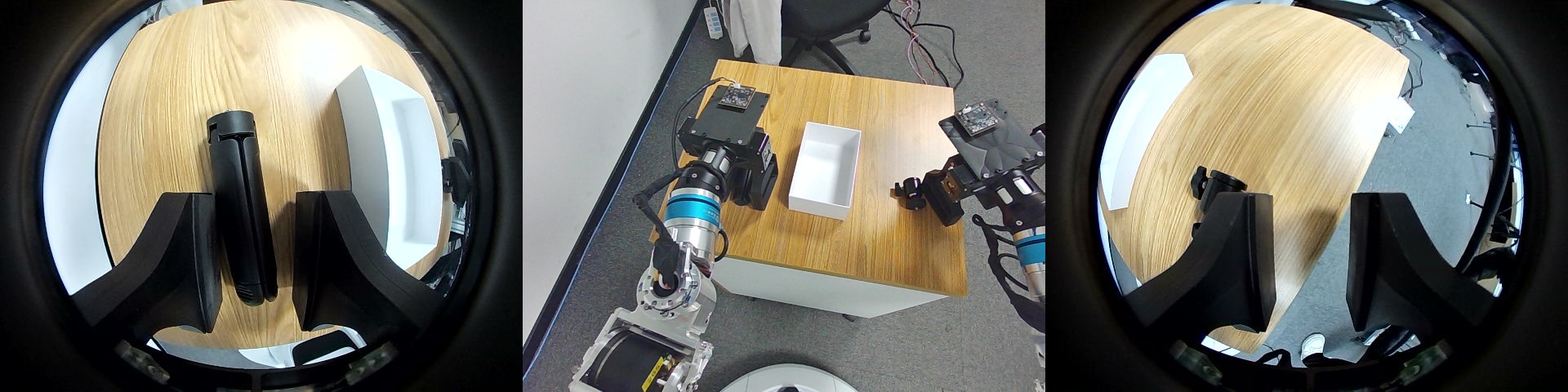}
    \caption{Remote Teleoperation Mode}
\end{subfigure}

\vspace{0.15cm}

\begin{subfigure}{\linewidth}
    \centering
    \includegraphics[width=\linewidth]{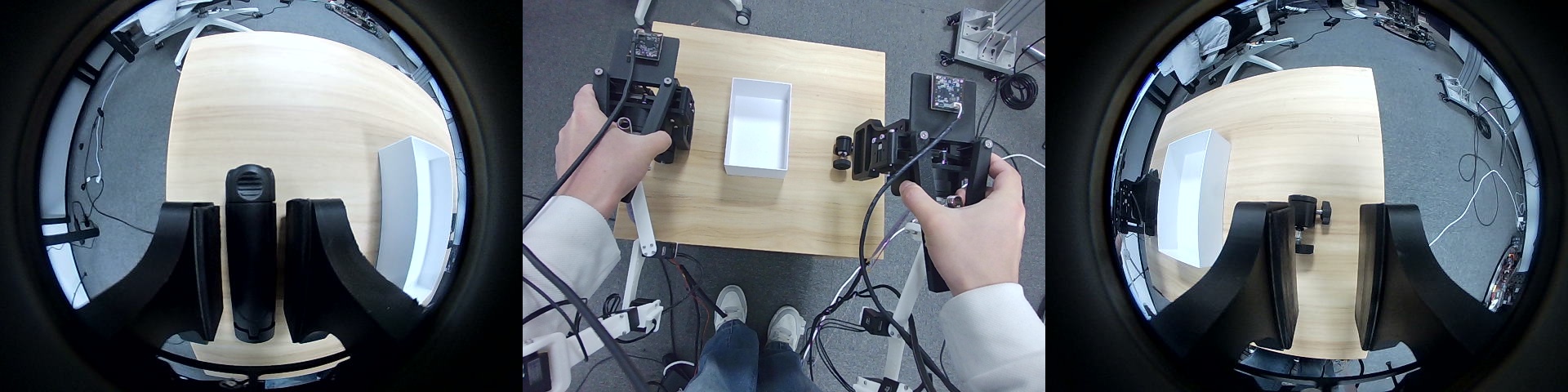}
    \caption{Active Demonstration Mode}
\end{subfigure}
\caption{\textbf{Bimodal Visuomotor Observations.} Chest (global) and wrist (localized) views during (a) Teleoperation and (b) Active Demonstration. Both modes record the same chest and left/right wrist views; the operator's arms and wearable in (b) are the main visual difference from deployment.
}
\label{fig:bimodal_data}
\end{minipage}
\end{figure}

SuperSuit uses mechanically aligned joint modules (Fig.~\ref{fig:cad_design}), so its arm joints correspond to the robot's one-to-one and no Cartesian inverse kinematics (IK) is solved.
A low-pass filter (LPF) smooths the recorded joint signals before action construction.

Matching joints avoids inverse kinematics, but it does not make the joint values exact: a zero-point calibration error shifts recorded joints by a constant, and gear backlash and structural compliance add errors that vary with load and pose.
An absolute target carries all of these into the supervised target.
To keep the constant shift out of it, we represent the arm targets in an $H$-step action chunk as $\Delta\mathbf{q}_{t,k}=\mathbf{q}^{cmd}_{t+k}-\mathbf{q}_t$, where $\mathbf{q}_t$ is the observed joint state, $\mathbf{q}^{cmd}_{t+k}$ is the recorded target command, and $k=0,\ldots,H-1$ indexes the chunk time offsets.
All targets in a chunk share the same observation $\mathbf{q}_t$ as their reference.
In teleoperation the observation is the robot's joint-encoder state, and in active demonstration it is the wearable arm's encoder state; both pass through the same per-joint zero and sign conversion (App.~\ref{app:arm_calibration}) and are stored in the robot joint convention, so the policy reads one state definition in either mode.

During inference, each predicted increment is converted to an absolute target $\widehat{\mathbf{q}}^{cmd}_{t,k}=\mathbf{q}_t+\Delta\mathbf{q}_{t,k}$ before execution.
Historical joint differences $(\mathbf{q}_t-\mathbf{q}_{t-1})$ provide observation velocity features.
The deployment runner executes a prefix of $M\leq H$ commands at a nominal 30~Hz and queries the policy again when its queue is empty; $M$ is configurable independently of $H$.
A constant offset cancels in $\Delta\mathbf{q}$ when the recorded command and the observed joint state share it (App.~\ref{app:action_schema}). An offset that differs between the two does not cancel, and neither do the time-varying compliance and backlash.

\subsection{LLM-Assisted Human-in-the-Loop Annotation Pipeline}
\label{subsec:data_pipeline}

During collection, the system logs visual observations, proprioception, the 22-D whole-body action, and an audio stream of operator narration.
The offline annotation workflow transcribes audio with Paraformer~\cite{paraformer} and uses Qwen3~\cite{qwen3} with kinematic events to propose subtask boundaries.
A human annotator reviews and corrects the proposed intervals and language labels.
Sec.~\ref{subsec:annotation_audit} evaluates the proposals by timestamp error against the corrected boundaries.

\section{Experiments}
\label{sec:experiments}
Our experiments address four questions: (1) \textbf{Data Collection Efficiency}: Does SuperSuit increase demonstration throughput over teleoperation baselines?
(2) \textbf{Data Utility}: How do active-only, mixed, and teleoperation datasets compare under matched policy training and evaluation?
(3) \textbf{Scaling}: How does success change as active data grow?
(4) \textbf{Design and annotation}: Which action, backbone, and annotation choices do the results depend on, and how accurate are the proposed subtask boundaries?

\subsection{Experimental Setup and Benchmark Tasks}
\label{subsec:task_setup}

\textbf{Hardware Platform.}
We evaluate SuperSuit on a custom 22-DoF wheeled bimanual humanoid: two 7-DoF arms with 1-DoF parallel grippers (16 DoF for manipulation), a 3-DoF active torso (lift, yaw, pitch), and a 3-DoF omnidirectional base ($v_x, v_y, \omega_z$).
Visual perception uses synchronized RGB from a chest-mounted ORBBEC Gemini 336L (global view) and bilateral wrist cameras (Fig.~\ref{fig:bimodal_data}).

\textbf{Policy Training.}
We train ACT~\cite{act} for $50{,}000$ steps with batch size $256$, AdamW, and learning rate $1 \times 10^{-4}$.
We use $\pi_{0.5}$~\cite{pi_zero_five} as the default backbone and evaluate subtask supervision with $\pi_{0.5}^{+}$, which adds an auxiliary autoregressive language head (loss weight $0.1$) for predicting the current subtask text.
Both $\pi_{0.5}$ variants are trained for $10{,}000$ steps with batch size $256$, AdamW, and learning rate $5 \times 10^{-5}$.
Unless stated otherwise, policy results use one training seed.
We evaluate three fixed checkpoints: 8k, 9k, and 10k steps for both $\pi_{0.5}$ variants, and 40k, 45k, and 50k steps for ACT.
Each checkpoint is evaluated with 20 real-robot rollouts per task.
Tables report the best of the three checkpoints for each condition; Wilson intervals in App.~\ref{app:statistics} describe rollout uncertainty at that checkpoint and do not capture seed or checkpoint-selection variation.

All methods use the same robot, camera observations, task randomization, success criteria, and rollout protocol.
The BRS-style policy baseline is an in-house implementation using the same exoskeleton for arm control and Joy-Cons for torso, base, and gripper commands; it is included as the controlled policy-data baseline because matched HOMIE, AirExo, or UMI implementations and datasets were unavailable.

\textbf{Task 1: Pick-and-Place.}
The robot navigates to a desk, retrieves a tripod and a connector, and places them into a box, which requires precise grasps at the end of a navigation segment.

\textbf{Task 2: Blocks Collection.}
Holding a receptacle in its right hand, the robot picks scattered blocks with its left hand over repeated pick cycles.

\textbf{Task 3: Crate Stacking.}
The robot lifts a crate with both arms, carries it to a target, and stacks it, holding contact through base and torso motion; a slip during transport loses the grasp.

\subsection{Data Collection Efficiency}
\label{subsec:data_efficiency}

We evaluate data collection efficiency across three modalities: a BRS-style baseline~\cite{brs}, SuperSuit Teleoperation, and SuperSuit Active Demonstration (Table~\ref{tab:throughput_transposed}).

\begin{table}[tb]
\centering
\footnotesize
\setlength{\tabcolsep}{2.5pt}
\captionsetup{justification=centering}
\caption{Collection throughput (episodes/hour). BRS-style denotes our in-house baseline.}
\label{tab:throughput_transposed}
\begin{tabular}{lccc}
\toprule
\textbf{Modality} & \textbf{Pick-Place} & \textbf{Blocks} & \textbf{Stacking} \\ \midrule
BRS-style~\cite{brs}        &  56.8 & 20.7  & 11.4 \\
SuperSuit (Teleop)    & 65.1  & 24.3  & 13.2 \\
SuperSuit (Active)    & \textbf{151.4} & \textbf{51.9} & \textbf{32.5} \\\bottomrule
\end{tabular}
\end{table}

SuperSuit (Teleop) records 15--17\% more valid episodes per measured collection hour than the BRS-style baseline.
A trajectory audit found pauses in about 6\% of BRS-style Stacking base/torso frames, which we did not isolate.
Active Demonstration records 2.5--2.9$\times$ as many episodes per measured collection hour as BRS-style (151.4, 51.9, and 32.5 episodes/hour).
The gain reflects the collection mode as a whole: no robot executes, resets, or recovers during active demonstration.

Each cell uses 30 valid demonstrations from one expert operator with approximately 1,000 hours of experience.
Episode time includes task resets, failed attempts and recovery, synchronous narration, and logging; it excludes one-time calibration, setup, and offline annotation (App.~\ref{app:throughput_formula}).

\subsection{Policy Performance and Effective Throughput}
\label{subsec:policy_performance}

\begin{table}[htbp]
\centering
\caption{Policy success rate and effective throughput under different training data compositions.}
\label{tab:policy_success}
\label{tab:effective_throughput}
\resizebox{\columnwidth}{!}{
\begin{tabular}{lcccccc}
\toprule
& \multicolumn{3}{c}{\textbf{Policy Success Rate (\%)}} & \multicolumn{3}{c}{\textbf{Effective Throughput (success/hour)}} \\
\cmidrule(lr){2-4} \cmidrule(lr){5-7}
\textbf{Training Data Composition}
& \textbf{Pick-Place}
& \textbf{Blocks}
& \textbf{Stacking}
& \textbf{Pick-Place}
& \textbf{Blocks}
& \textbf{Stacking} \\
\midrule
Teleop 10 / Active 0   & 10\% & 0\% & 0\% & 4.2 & 0.0 & 0.0 \\
Teleop 110 / Active 0  & 85\% & 65\% & 40\% & 41.9 & 11.4 & 3.8 \\
Teleop 10 / Active 100 & 85\% & 60\% & 40\% & 88.8 & 23.3 & 9.5 \\
\bottomrule
\end{tabular}
}
\end{table}

We compare $\pi_{0.5}$ data compositions at a fixed 110-episode dataset size (Table~\ref{tab:policy_success}).
Active-only data performs well below teleoperation-only data (Table~\ref{tab:app_principal_ci}), while keeping 10 robot-grounded episodes and replacing the other 100 with active demonstrations comes within 0, 1, and 0 successes out of 20 of teleoperation-only data (at most two at any matched checkpoint). These 10 episodes may do two jobs at once: they carry the shared action contract, and they are the only robot-arm visual and contact data in the mixture. This study does not separate the two, so the benefit of the hybrid set should not be read as coming from the action schema alone.

Effective throughput is the expected number of successful autonomous completions per evaluation hour: success rate times evaluation cycles per hour, where the cycle time counts failed and timed-out trials plus reset and recovery (App.~\ref{app:throughput_formula}).
Unlike collection throughput, it is a property of the trained policy.
The BRS-style and SuperSuit teleoperation rows, trained and evaluated with the same robot, observations, recipe, and protocol, differ by at most one success out of 20 per task (Table~\ref{tab:app_principal_ci}).

\subsection{Scalability Analysis}
\label{subsec:scaling}

\begin{table}[htbp]
\centering
\begin{minipage}{0.48\textwidth}
\centering
\footnotesize
\setlength{\tabcolsep}{3pt}
\caption{Success rate (\%) as active data grow with a fixed 10-episode teleoperation seed.}
\label{tab:scaling}
\begin{tabular}{lccc}
\toprule
\textbf{Teleop / Active} & \textbf{Pick-Place} & \textbf{Blocks} & \textbf{Stacking} \\
\midrule
10 / 0   & 10\% & 0\% & 0\% \\
10 / 50  & 65\% & 25\% & 15\% \\
10 / 100 & 85\% & 60\% & 40\% \\
10 / 200 & 100\% & 60\% & 55\% \\
10 / 400 & 100\% & 75\% & 65\% \\
\bottomrule
\end{tabular}
\end{minipage}
\hfill
\begin{minipage}{0.48\textwidth}
\centering
\footnotesize
\setlength{\tabcolsep}{3pt}
\caption{Ablation on action formulation and velocity deadband on Crate Stacking.}
\label{tab:ablation_delta}
\begin{tabular}{lc}
\toprule
\textbf{Configuration} & \textbf{Success Rate} \\
\midrule
$\Delta q$ w/ Deadband (Ours) & \textbf{40\%} \\
$\Delta q$ w/o Deadband & 30\% \\
Absolute Joint ($q$) & 5\% \\
\bottomrule
\end{tabular}
\end{minipage}
\end{table}

With teleoperation fixed at 10 episodes, the best-checkpoint success rate is non-decreasing as active demonstrations grow from 0 to 400 on each of the three tasks (Table~\ref{tab:scaling}).
A matched teleoperation-only curve is also available for Crate Stacking: 0/20, 4/20, 8/20, 12/20, and 15/20 successes at 10, 60, 110, 210, and 410 demonstrations.
At matched sizes the hybrid curve stays within two successes of it (3/8/11/13 vs.\ 4/8/12/15 out of 20 at 60--410 episodes), and at the Crate Stacking rates in Table~\ref{tab:throughput_transposed} the 410-episode hybrid set takes about 13 collection hours, of which under one requires the robot, versus about 31 hours for teleoperation.
No corresponding teleoperation scaling curves were collected for Pick-and-Place or Blocks Collection.

\subsection{Ablation Study: Action Formulation and Velocity Deadband}
\label{subsec:ablation_delta}

To isolate action formulation and deadband, we selectively disable each component under Teleop 10 / Active 100 (Table~\ref{tab:ablation_delta}).
At the best checkpoint of each configuration, $\Delta q$ with deadband gives 8/20 successes versus 1/20 for absolute joints and 6/20 without the deadband.
Across the three checkpoints, $\Delta q$ plus deadband exceeds absolute targets (8/8/7 vs. 1/0/0); paired exact McNemar tests with Holm correction give adjusted $p\leq0.0391$.
Deadband-on success is higher at every checkpoint (8/8/7 vs. 4/6/4), but this comparison is not statistically resolved.
A plausible mechanism is that, without the deadband, operator sway is recorded as base motion and replayed while both grippers hold the crate, shifting it relative to the grippers (App.~\ref{app:deadband_failure}).

\subsection{Language Annotation Audit}
\label{subsec:annotation_audit}

On 100 Crate Stacking demonstrations, the human-in-the-loop annotation workflow required a mean of 15 s per episode for correction, compared with 37 s for fully manual annotation (59\% lower correction time).
There are four internal boundaries per five-subtask episode, or 400 boundaries total.
Of the model-proposed timestamps, 327/400 were within $\pm1$ s of the final human-corrected boundary (81.8\%, 95\% Wilson CI [77.7,85.2]\%).
This measures boundary timing only, not missed or extra proposals.

\subsection{Ablation Study: Policy Backbone and Subtask Annotations}
\label{subsec:subtask_ablation}

We evaluate two orthogonal factors on the same hybrid dataset (Teleop 10 / Active 100, Table~\ref{tab:subtask_ablation}): policy backbone choice and subtask annotation supervision.
ACT uses no language; $\pi_{0.5}$ receives the task instruction; only $\pi_{0.5}^{+}$ receives verified subtask labels.

\begin{table}[H]
\centering
\footnotesize
\setlength{\tabcolsep}{3pt}
\caption{Ablation on policy backbone and subtask annotations.}
\label{tab:subtask_ablation}
\begin{tabular}{lccc}
\toprule
\textbf{Architecture} & \textbf{Pick-Place} & \textbf{Blocks} & \textbf{Stacking} \\
\midrule
ACT~\cite{act}   & 70\% & 30\% & 20\% \\
$\pi_{0.5}$~\cite{pi_zero_five}   & 85\% & 60\% & 40\% \\
$\pi_{0.5}^{+}$   & 85\% & 65\% & 50\% \\
\bottomrule
\end{tabular}
\end{table}

\textbf{Backbone capacity and pretraining.}
ACT reaches 70\% / 30\% / 20\% and $\pi_{0.5}$ 85\% / 60\% / 40\%, gaps of 3, 6, and 4 out of 20. The gaps are not ordered by task length, and backbone, pretraining, and recipe all change together, so we read no trend into them.

\textbf{Subtask annotation supervision.}
Adding verified subtask labels ($\pi_{0.5}^{+}$) gives 85\% / 65\% / 50\%, 0, 1, and 2 more successes out of 20. The direction is consistent, but at 20 rollouts no difference is resolvable, so this ablation leaves the benefit unestablished.

\section{Conclusion}
SuperSuit turns one wearable into both the controller and the data source for a wheeled bimanual robot.
Because the arms align with the robot's joint for joint and a single tracker stands in for the torso and the base, one policy trains on what the operator does whether or not the robot is present.
That is what no existing interface provides: keeping the robot in the loop spends robot hours on every episode, and recording human motion without it inserts a mapping between what the operator did and what the policy is asked to reproduce, so the two cannot train the same policy.
Both roles here emit the same 22-D supervision instead, and the effect is to move where the cost of data collection sits: active collection yields far more episodes per hour than teleoperation, and a dataset built mostly from active demonstrations matches the policy quality of a fully teleoperated one while needing the robot for only a small share of its episodes.
Active-only data falls short, so a small robot-grounded seed is what anchors the mixture, and the same recordings yield verified subtask labels at a fraction of the manual correction time.

\section{Limitations}
Our evaluation uses one wheeled bimanual robot, one expert operator, and one training seed per policy configuration; operators and workspaces were not varied.
Success rates come from the best of three checkpoints at 20 rollouts each, so close counts do not establish equivalence.
Lighthouse tracking requires an instrumented workspace and clear line of sight, and the torso reference assumes the suit stays rigidly coupled to the operator's torso.
On-device SLAM for velocity-level base control remains untested.
Collection rates exclude one-time setup and offline annotation.

Direct joint correspondence assumes similar wearable and robot kinematics.
Adapting the suit to a second 7-DoF platform took about 10 person-hours but does not establish cross-embodiment policy transfer; dissimilar arms require retargeting.
Experiments use flat-paddle grippers.
The end effector is a replaceable module on the wrist interface, but dexterous tasks and the two-finger gripper pair (App.~\ref{app:end_effector}) were not evaluated.
The printed suit may exhibit load-dependent compliance and backlash.
The visual, proprioceptive, and contact context of robot-free collection does not match deployment, and we did not isolate which difference accounts for the active-only gap.
Annotation audits omit missed/extra boundaries, and policy results do not establish a language-supervision benefit.

\bibliography{references}

\clearpage

\appendix
\section*{Appendix}
\section{System Implementation Details}
\label{app:system}

This appendix provides implementation and evaluation details that complement the main paper, including hardware composition, calibration, data schemas, policy training, rollout uncertainty, annotation timing, throughput accounting, and failure analysis.

\subsection{Robot Platform}
\label{app:robot_platform}

All real-world experiments use the custom wheeled bimanual platform described in Sec.~\ref{subsec:task_setup}.
The platform contains two 7-DoF arms, two 1-DoF parallel grippers, a 3-DoF active torso, and a 3-DoF omnidirectional base.
We treat the torso and base as part of the whole-body action space because long-horizon mobile manipulation requires coordinated navigation, upper-body positioning, and bimanual manipulation.

\begin{table}[ht]
\centering
\footnotesize
\setlength{\tabcolsep}{4pt}
\caption{Robot hardware summary.}
\label{tab:app_robot_hardware}
\resizebox{\linewidth}{!}{
\begin{tabular}{lll}
\toprule
\textbf{Component} & \textbf{Specification} & \textbf{Role in SuperSuit experiments} \\
\midrule
Left/right arms & 7 DoF per arm & Bimanual manipulation and object transport \\
Left/right grippers & 1 DoF parallel gripper per arm & Binary or continuous grasping command \\
Active torso & Lift, yaw, and pitch & Aligns upper-body workspace with the task scene \\
Mobile base & Omnidirectional $v_x$, $v_y$, $\omega_z$ command & Planar locomotion during long-horizon tasks \\
Global camera & Chest-mounted RGB camera, $640\times512$ at 30 fps & Global scene observation \\
Wrist cameras & Left/right fisheye RGB cameras, $640\times512$ at 30 fps & Local manipulation observation near each gripper \\
\bottomrule
\end{tabular}
}
\end{table}

\subsection{End-Effector Variants}
\label{app:end_effector}

The platform used in all reported experiments carries a 1-DoF parallel gripper with flat paddle fingers (Table~\ref{tab:app_robot_hardware}).
The flat geometry is sufficient for the box-like objects in the three evaluated tasks, but it cannot form enclosures around irregular or thin objects, which bounds the manipulation primitive to planar contact.

\begin{figure}[H]
\centering
\begin{subfigure}{0.34\linewidth}
    \centering
    \includegraphics[width=\linewidth]{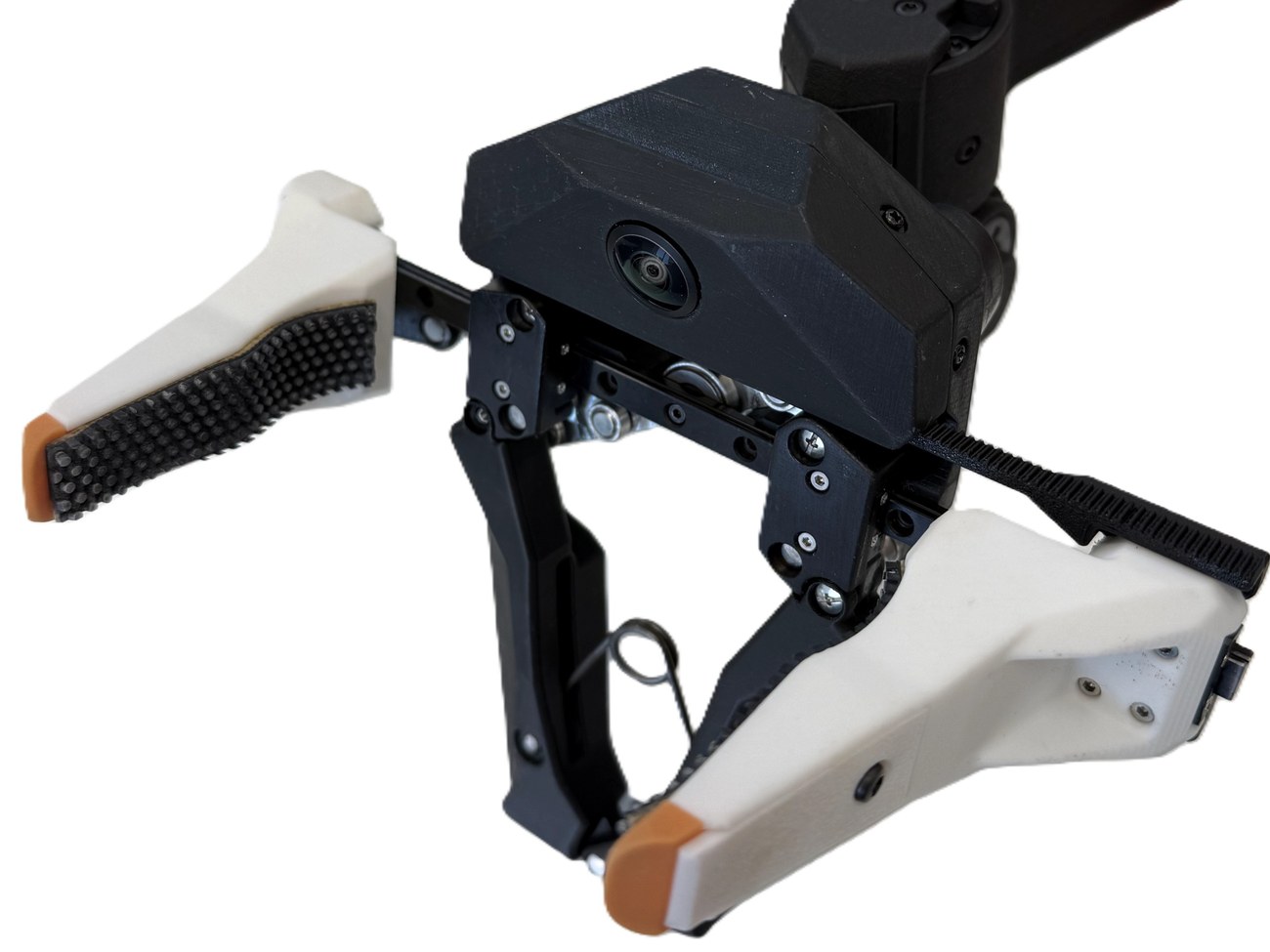}
    \caption{SuperSuit-side gripper}
    \label{fig:end_effector_wearable}
\end{subfigure}\hspace{0.04\linewidth}
\begin{subfigure}{0.34\linewidth}
    \centering
    \includegraphics[width=\linewidth]{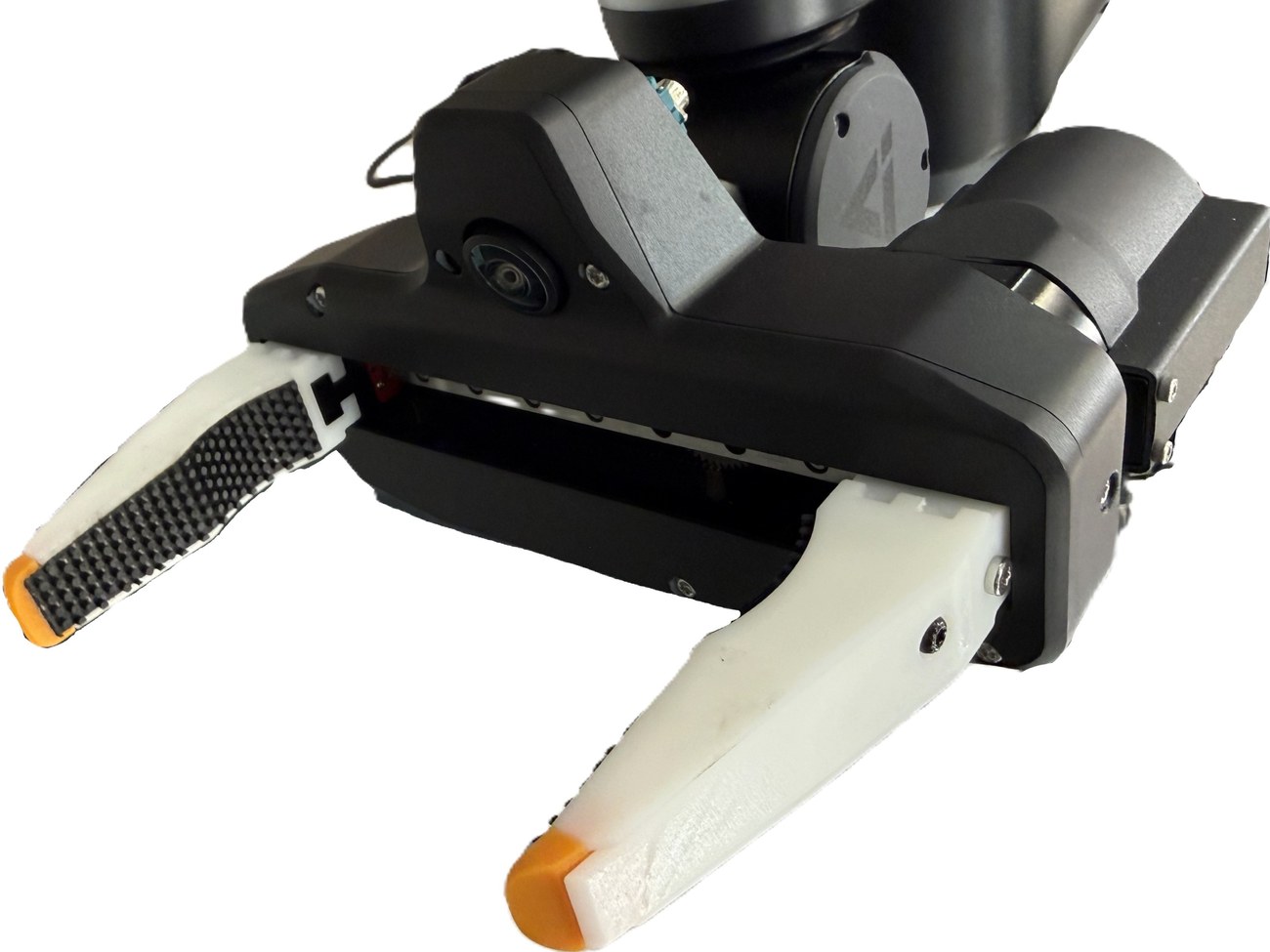}
    \caption{Robot-side gripper}
    \label{fig:end_effector_robot}
\end{subfigure}
\caption{\textbf{Two-finger gripper pair.} The SuperSuit-side gripper used for data collection and the matching robot-side gripper, which share finger geometry and wrist-camera placement. }
\label{fig:end_effector}
\end{figure}

A matched pair of two-finger grippers is available (Fig.~\ref{fig:end_effector}), with matching finger geometry and wrist-camera placement on the SuperSuit and robot sides.
The end effector is a replaceable module: the pair mounts on the same wrist interface and uses the same normalized scalar command $g_t \in [0,1]$, so the arms, gripper channel, and 22-D action are unchanged.
The pair was not used for any data or policy result reported here; its evaluation remains future work.

\subsection{Wearable Interface}
\label{app:wearable_interface}

The SuperSuit wearable interface (Table~\ref{tab:app_wearable_hardware}) is composed of an upper-body isomorphic arm mechanism, a spatial tracker on a back-fixed bracket, and an integrated microphone.
The arm mechanism mirrors the target arm joint structure, allowing joint-space commands to be recorded without inverse kinematics when the embodiment is matched.
The tracker sits above the operator's head on a bracket fixed to the back of the suit, where it avoids hand and arm occlusion without restricting head movement.
The microphone records in-situ narration, which is aligned with physical subtasks during offline annotation.

\begin{table}[ht]
\centering
\footnotesize
\setlength{\tabcolsep}{4pt}
\caption{Wearable hardware summary.}
\label{tab:app_wearable_hardware}
\resizebox{\linewidth}{!}{
\begin{tabular}{lll}
\toprule
\textbf{Subsystem} & \textbf{Implementation} & \textbf{Notes} \\
\midrule
Arm interface & Isomorphic left/right wearable arms & Robot-aligned joints \\
Joint modules & Dynamixel XL330-M288-T servos & Joint logging \\
Gripper interface & Robot-matched gripper handles & Matched appearance \\
Spatial tracker & HTC Vive Tracker on a back-fixed bracket, with HTC Base Station 2.0 & Base/torso tracking \\
Audio input & Headset microphone & Narration capture \\
Wearable mass & 3.2 kg & Full wearable system \\
\bottomrule
\end{tabular}
}
\end{table}

\subsection{Collection Modes}
\label{app:modes}

SuperSuit supports two data-collection modes under a shared representation.

\paragraph{Remote teleoperation.}
The operator wears SuperSuit while controlling the physical robot in real time.
Visual feedback is provided by the robot cameras, and the robot executes the mapped torso, base, arm, and gripper commands during collection.
This mode preserves robot embodiment in the visual stream and is used to collect a small amount of robot-grounded seed data.

\paragraph{Active demonstration.}
The operator performs the task directly in the workspace while wearing SuperSuit.
The robot is not physically executing the motion during recording, which removes robot-in-the-loop latency and hardware reset time.
Because the same exoskeleton, tracker, action formulation, and language pipeline are used, active demonstrations share the same action schema as teleoperated demonstrations.
Active demonstrations contain wearable rather than robot arms and also differ in proprioceptive and contact context, creating the domain gap discussed in the limitations.

\section{Calibration and Signal Processing}
\label{app:calibration}

\subsection{Coordinate Frames}
\label{app:frames}

We use the following coordinate frames throughout the system:
\begin{itemize}
    \item $\mathcal{F}_W$: tracking world frame defined by the Lighthouse system.
    \item $\mathcal{F}_H$: tracker frame.
    \item $\mathcal{F}_T$: tracker-derived torso reference frame, using the robot torso pitch-link convention.
    \item $\mathcal{F}_0$: fixed initialization frame, coincident with $\mathcal{F}_T$ at time $t_0$.
    \item $\mathcal{F}_B$: mobile-base frame.
    \item $\mathcal{F}_{A}^{L}$ and $\mathcal{F}_{A}^{R}$: left and right arm joint frames.
\end{itemize}

We use homogeneous column vectors and define ${}^{A}\!\mathbf{T}_{B}\in SE(3)$ by ${}^{A}\!\mathbf{p}={}^{A}\!\mathbf{T}_{B}{}^{B}\!\mathbf{p}$. Thus, ${}^{A}\!\mathbf{T}_{B}$ maps coordinates from frame $B$ to frame $A$. The raw tracker pose is $\mathbf{T}_h(t)={}^{W}\!\mathbf{T}_{H}(t)$, and the fixed mounting transform is $\mathbf{T}_{cal}={}^{H}\!\mathbf{T}_{T}$. It is applied on the right:
\begin{equation}
    {}^{W}\!\mathbf{T}_{T}(t) = {}^{W}\!\mathbf{T}_{H}(t){}^{H}\!\mathbf{T}_{T}=\mathbf{T}_h(t)\mathbf{T}_{cal}.
\end{equation}
At initialization, we record ${}^{W}\!\mathbf{T}_{T}(t_0)$. The pose used for command generation is expressed in the fixed initial frame:
\begin{equation}
    {}^{0}\!\mathbf{T}_{T}(t)=\left[{}^{W}\!\mathbf{T}_{T}(t_0)\right]^{-1}{}^{W}\!\mathbf{T}_{T}(t).
\end{equation}
We decompose this relative pose into position and Euler components $(x,y,z,\psi,\theta,\phi)$. The single yaw signal is allocated by the locomotion detector described in Sec.~\ref{subsec:locomotion}: its residual relative to the commanded base heading becomes the torso-yaw target when stationary and a heading-error feedback on the base when locomotion is active, so the torso-yaw target returns to zero during locomotion without losing heading. The base command $(v_x,v_y)$ is differentiated in the fixed initialization frame $\mathcal{F}_0$ and then rotated into the torso reference $\mathcal{F}_T$ before it is issued to the chassis. The torso-yaw target is zero whenever the locomotion detector is active (Sec.~\ref{subsec:locomotion}), which is when the base command is nonzero, so $\mathcal{F}_T$ and the base frame $\mathcal{F}_B$ share their $x$ and $y$ axes and no further rotation is required.

\subsection{Tracker-to-Torso Calibration}
\label{app:tracker_calibration}

The mounting transform $\mathbf{T}_{cal}={}^{H}\!\mathbf{T}_{T}$ defines a tracker-derived torso reference. The implementation composes an axis-alignment rotation with a translation:
\begin{equation}
    \mathbf{T}_{cal}=\begin{bmatrix}\mathbf{R}_{\mathrm{align}} & \mathbf{R}_{\mathrm{align}}\mathbf{d} \\\mathbf{0}^{\top} & 1\end{bmatrix},\quad \mathbf{R}_{\mathrm{align}}=\mathbf{R}_z(-\pi/2)\mathbf{R}_x(\pi),\quad \mathbf{d}=[0.25,0,-0.55]^\top\mathrm{m}.
\end{equation}
The vector $\mathbf{d}$ is expressed in the aligned local axes, so the torso-reference origin in tracker coordinates is $\mathbf{R}_{\mathrm{align}}\mathbf{d}$. The transform is reused for the same mounting geometry, while the initial reference pose is recorded at each setup. The bracket is fixed to the suit rather than to the head, so $\mathcal{F}_T$ follows the operator's back and head motion does not enter the mapping.
Interpreting $\mathcal{F}_T$ as the physical torso pose therefore assumes only that the suit stays rigidly coupled to the torso.

\subsection{Arm and Gripper Calibration}
\label{app:arm_calibration}

The wearable arm records absolute joint positions $\tilde{\mathbf{q}}_t$, while the robot controller receives commands in the robot joint convention. We maintain a per-joint zero offset $\mathbf{b}$ and sign vector $\mathbf{s}$:
\begin{equation}
    \mathbf{q}_t = \mathbf{s}\odot(\tilde{\mathbf{q}}_t-\mathbf{b}),
\end{equation}
where $\odot$ denotes element-wise multiplication. The same conversion is applied in both teleoperation and active demonstration so that the two modalities remain compatible. Gripper commands are normalized to a common scalar range:
\begin{equation}
    g_t = \frac{\tilde{g}_t-g_{\min}}{g_{\max}-g_{\min}},
\end{equation}
with $g_t=0$ and $g_t=1$ corresponding to fully closed and fully open states, respectively.

\subsection{Temporal Filtering}
\label{app:filtering}

Tracker position and Euler components are temporally filtered after the mounting-transform and initial-reference conversion; joint signals are filtered before action construction. We use an exponential low-pass filter:
\begin{equation}
    \bar{\mathbf{x}}_t = (1-\alpha)\mathbf{x}_t + \alpha\bar{\mathbf{x}}_{t-1},
\end{equation}
where $\mathbf{x}_t$ denotes either relative tracker pose components or joint positions. We set $\alpha=0.95$ as the history weight, so the previous filtered state contributes 95\% and the new measurement contributes 5\%, providing low-pass smoothing of high-frequency measurement noise. The velocity deadband separately suppresses low-amplitude base commands. Angular components are unwrapped before filtering and wrapped back to the principal interval afterward, avoiding artificial jumps at the Euler-angle branch cut.

\subsection{Velocity Deadband}
\label{app:deadband}

The purpose of this module is to prevent small involuntary torso motions from being recorded as base commands. After differentiating the filtered tracker trajectory, low-amplitude velocities are suppressed by a dimensional deadband, while deliberate locomotion commands are scaled and clipped before being sent to the base. We implement the deadband as
\begin{equation}
\mathcal{D}(v,\epsilon)=
\begin{cases}
0, & |v|\leq \epsilon,\\
v, & |v| > \epsilon.
\end{cases}
\label{eq:app_deadband}
\end{equation}
The base command is then formed by scaling and componentwise clipping:
\begin{equation}
\mathbf{u}^{raw}_t =
\begin{bmatrix}
K_x \mathcal{D}(v_{h,x}(t),\epsilon_x)\\
K_y \mathcal{D}(v_{h,y}(t),\epsilon_y)\\
K_\omega \mathcal{D}(\omega_h(t),\epsilon_\omega)
\end{bmatrix}.
\end{equation}
For $\mathbf{u}_{max}=[0.35,0.35,0.8]^T$, clipping is defined componentwise by
\begin{equation}
\operatorname{clip}(\mathbf{u},\mathbf{u}_{max})_i
=\operatorname{sign}(u_i)\min(|u_i|,u_{max,i}),
\qquad \mathbf{u}^{base}_t=\operatorname{clip}(\mathbf{u}^{raw}_t,\mathbf{u}_{max}).
\end{equation}
The three entries are interpreted as $(v_x,v_y,\omega_z)$ in $\mathcal{F}_T$, which shares its $x$ and $y$ axes with the base frame while the torso-yaw target is zero. Table~\ref{tab:app_base_params} lists the gains, deadband thresholds, and clipping limits.
The policy retains the full 3-DoF base command, including forward velocity, lateral velocity, and yaw rate.

\begin{table}[ht]
\centering
\footnotesize
\caption{Base-command parameters. Values are conservative engineering defaults for indoor mobile manipulation and can be adjusted for a specific base.}
\label{tab:app_base_params}
\begin{tabular}{lcc}
\toprule
\textbf{Parameter} & \textbf{Meaning} & \textbf{Value} \\
\midrule
$K_x$ & Forward velocity scale & $1.0$ \\
$K_y$ & Lateral velocity scale & $1.0$ \\
$K_\omega$ & Yaw velocity scale & $1.0$ \\
$\epsilon_x$ & Forward deadband threshold & $0.03$ m/s \\
$\epsilon_y$ & Lateral deadband threshold & $0.03$ m/s \\
$\epsilon_\omega$ & Yaw deadband threshold & $0.05$ rad/s \\
Command clipping & Maximum allowed base velocity & $|v_x|,|v_y|\leq0.35$ m/s; $|\omega_z|\leq0.8$ rad/s \\
\bottomrule
\end{tabular}
\end{table}

\section{Action and Observation Representation}
\label{app:data_representation}

\subsection{Whole-Body State and Action Schema}
\label{app:action_schema}

For learning, SuperSuit stores time-aligned proprioceptive states and whole-body action commands. The raw proprioceptive vector is ordered as
\begin{equation}
\mathbf{p}_t =
\left[
\mathbf{q}^{L}_{t},
g^L_t,
\mathbf{q}^{R}_{t},
g^R_t,
\mathbf{q}^{torso}_{t}
\right],
\end{equation}
where $\mathbf{q}^{L}_{t},\mathbf{q}^{R}_{t}\in\mathbb{R}^{7}$ are the left and right arm joint states, $g^L_t,g^R_t\in\mathbb{R}$ are the gripper states, and $\mathbf{q}^{torso}_{t}\in\mathbb{R}^{3}$ contains torso lift, yaw, and pitch. This raw proprioceptive vector is 19-D.

Consistent with Sec.~\ref{subsec:isomorphism}, the policy observation augments the current proprioceptive vector with finite-difference proprioceptive velocities:
\begin{equation}
\mathbf{s}_t =
\left[
\mathbf{p}_t,
\dot{\mathbf{p}}_t
\right],
\quad
\dot{\mathbf{p}}_t =
\frac{\mathbf{p}_t-\mathbf{p}_{t-1}}{\Delta t}.
\end{equation}
The policy input is therefore the 38-D concatenation of these two blocks, both computed during preprocessing.

For an observation at step $t$, the policy predicts an $H$-step chunk $\{\mathbf{a}_{t,k}\}_{k=0}^{H-1}$, ordered as
\begin{equation}
\mathbf{a}_{t,k} =
\left[
\Delta\mathbf{q}^{L}_{t,k},
g^L_{t+k},
\Delta\mathbf{q}^{R}_{t,k},
g^R_{t+k},
\mathbf{q}^{torso,cmd}_{t+k},
\mathbf{u}^{base}_{t+k}
\right],
\label{eq:app_action}
\end{equation}
where the two arm increments are each 7-D, the gripper commands are scalar, the torso target is a 3-D non-incremental joint command whose lift and pitch come from the $\mathcal{F}_0$-relative pose and whose yaw comes from the allocation in Eq.~\eqref{eq:locomotion_gate}, and the base velocity command is 3-D in the torso reference $\mathcal{F}_T$, after rotation from $\mathcal{F}_0$. Each 22-D action uses commands from the same synchronized row $t+k$. Only the arm targets are converted to increments.

The arm action is defined as
\begin{equation}
    \Delta\mathbf{q}_{t,k} = \mathbf{q}^{cmd}_{t+k}-\mathbf{q}_t,
\end{equation}
where $\mathbf{q}_t$ is the observed arm state and $k=0,\ldots,H-1$ is a time offset corresponding to $k/30$ seconds at the 30~Hz sampling rate. The $k=0$ increment need not be zero because the recorded command can differ from the observed state. Chunk lengths $H$ are listed in Table~\ref{tab:app_policy_hyperparams}.

At inference, all arm increments are converted to absolute targets using the same observed state, $\widehat{\mathbf{q}}^{cmd}_{t,k}=\mathbf{q}_t+\Delta\mathbf{q}_{t,k}$.
These targets are executed without accumulating increments or changing their reference within a chunk.
The deployment runner consumes the first $M\leq H$ commands at a nominal 30~Hz and queries the policy again with the latest observation when its queue is empty.
$M$ is configurable independently of $H$.

The $\Delta q$ target is invariant to a constant offset shared by the recorded command and the observed joint state after both signals are expressed in the same joint-coordinate convention. Suppose a constant residual offset $\boldsymbol{\delta}$ remains after the conversion in App.~\ref{app:arm_calibration}, for example from an inaccurate zero offset $\mathbf{b}$:
\begin{equation}
    \mathbf{q}'_t = \mathbf{q}_t + \boldsymbol{\delta}.
\end{equation}
If the recorded target commands contain the same constant offset, $\mathbf{q}'^{\,cmd}_{t+k}=\mathbf{q}^{cmd}_{t+k}+\boldsymbol{\delta}$, an absolute action inherits this offset. In contrast, the forward-looking delta action satisfies
\begin{equation}
    \mathbf{q}'^{\,cmd}_{t+k}-\mathbf{q}'_{t}
    =
    (\mathbf{q}^{cmd}_{t+k}+\boldsymbol{\delta})-(\mathbf{q}_{t}+\boldsymbol{\delta})
    =
    \mathbf{q}^{cmd}_{t+k}-\mathbf{q}_{t}.
\end{equation}
Thus, a shared fixed offset in the matched joint convention is removed from the supervised target. An offset that differs between the command and the observed state is not removed, and neither are dynamic errors such as load-dependent compliance or backlash hysteresis.

\subsection{Collected Data Schema}
\label{app:observation_schema}

Each episode contains synchronized visual, proprioceptive, command, audio, and metadata streams.
The raw collected data record is shown in Table~\ref{tab:app_observation_schema}.

\begin{table}[ht]
\centering
\footnotesize
\setlength{\tabcolsep}{3.5pt}
\caption{Raw collected data record.}
\label{tab:app_observation_schema}
\resizebox{\linewidth}{!}{
\begin{tabular}{lll}
\toprule
\textbf{Field} & \textbf{Content} & \textbf{Dimension or format} \\
\midrule
Chest RGB & Global scene image & $640\times512\times3$ at 30 fps \\
Left wrist RGB & Left fisheye local manipulation image & $640\times512\times3$ at 30 fps \\
Right wrist RGB & Right fisheye local manipulation image & $640\times512\times3$ at 30 fps \\
Raw proprioception & $[\text{left-arm},\text{left-gripper},\text{right-arm},\text{right-gripper},\text{torso}]$ & 19-D vector \\
Action & $[\text{left-arm},\text{left-gripper},\text{right-arm},\text{right-gripper},\text{torso},\text{base}]$ & 22-D vector \\
Audio & Headset microphone stream & Timestamped WAV file \\
Metadata & Mode, task, episode id, timestamps & JSON or equivalent structured record \\
\bottomrule
\end{tabular}
}
\end{table}

\subsection{Time Synchronization and Episode Segmentation}
\label{app:synchronization}

All streams are timestamped against a shared host clock before being written to disk.
Camera frames, wearable encoder readings, tracker poses, robot proprioception, base commands, gripper states, and audio packets are aligned primarily by nearest-neighbor timestamp matching.
The policy frequency is 30 Hz.
Episodes begin when the operator starts a task-level recording and terminate when the task is completed, a safety stop is triggered, or the maximum horizon is reached.
During preprocessing, static segments without meaningful robot or wearable motion are removed, while short pauses within an ongoing task are retained.

\section{Language Annotation Pipeline}
\label{app:annotation}

\subsection{Timestamped Transcription}
\label{app:transcription}

During each demonstration, the operator narrates the current subtask in natural language. The audio stream is transcribed with Paraformer~\cite{paraformer}, producing a sequence of timestamped text segments:
\begin{equation}
    \mathcal{S} = \{(r_i^{start}, r_i^{end}, y_i)\}_{i=1}^{N_s}.
\end{equation}
The transcription is not assumed to be perfectly synchronized with physical execution, since operators often describe an action before or after the corresponding motion.

\subsection{Kinematic Breakpoint Proposal}
\label{app:breakpoints}

\begin{figure}[H]
    \centering
    \includegraphics[width=0.82\linewidth]{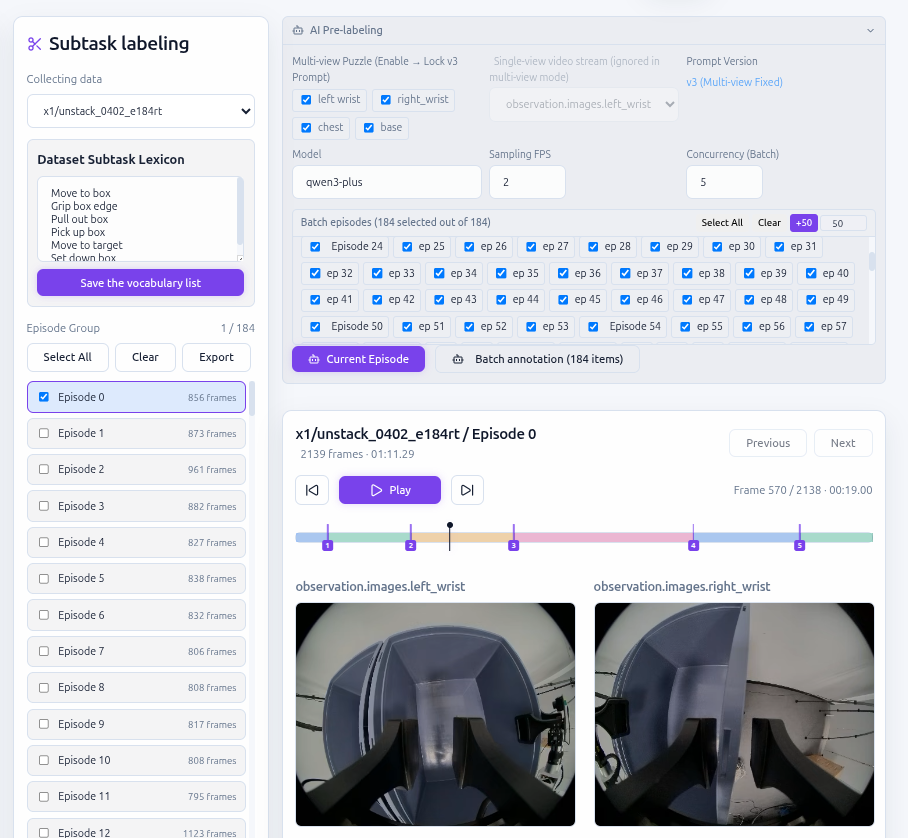}
    \caption{\textbf{Subtask annotation interface.}
    The annotator presents the task-level subtask vocabulary, batch pre-labeling controls, synchronized playback, timeline markers, and multi-view robot observations. Human annotators use this interface to verify model-proposed subtask boundaries and correct the language label associated with each interval.}
    \label{fig:app_subtask_annotator}
\end{figure}

To align speech with action, we compute candidate physical breakpoints from the whole-body trajectory using changes in arm, base, torso, and gripper motion.
These cues capture pauses, locomotion changes, grasp transitions, arm-activity switches, and large torso adjustments.
Qwen3~\cite{qwen3} receives the timestamped transcript and a compact summary of these kinematic events, then proposes subtask intervals.
Human annotators verify and correct the resulting intervals in the interface of Fig.~\ref{fig:app_subtask_annotator}, because narration boundaries and physical state transitions are often close but not identical.

\subsection{Annotation Output}
\label{app:annotation_output}

The verified annotation for each episode is stored as a list of subtask records:
\begin{equation}
    \mathcal{A} = \{(t_j^{start}, t_j^{end}, \ell_j)\}_{j=1}^{N_a},
\end{equation}
where $\ell_j$ is the verified language instruction for the subtask. These labels are used by $\pi_{0.5}^{+}$ as auxiliary subtask supervision.

\section{Training Details}
\label{app:training}

\subsection{Dataset Organization}
\label{app:dataset_compositions}

The dataset compositions reported in the main paper are constructed independently for each benchmark task.
Teleoperation and active demonstrations are stored with the same state/action schema described in Sec.~\ref{app:action_schema}, so hybrid datasets are formed by concatenating episodes from the two collection modes without modality-specific action conversion.
For the scalability study, the teleoperation subset is kept fixed while active episodes are appended in increasing amounts, so that the teleoperation seed set is identical across the curve.

\subsection{Policy Hyperparameters}
\label{app:policy_hyperparams}

Table~\ref{tab:app_policy_hyperparams} summarizes the policy hyperparameters used in the main experiments.
All methods use the same synchronized observation and action representation unless otherwise stated.

\begin{table}[ht]
\centering
\footnotesize
\setlength{\tabcolsep}{4pt}
\caption{Policy training hyperparameters.}
\label{tab:app_policy_hyperparams}
\resizebox{\linewidth}{!}{
\begin{tabular}{llll}
\toprule
\textbf{Parameter} & \textbf{ACT} & \textbf{$\pi_{0.5}$} & \textbf{$\pi_{0.5}^{+}$} \\
\midrule
Training steps & $50{,}000$ & $10{,}000$ & $10{,}000$ \\
Batch size & $256$ & $256$ & $256$ \\
Optimizer & AdamW & AdamW & AdamW \\
Learning rate & $1\times10^{-4}$ & $5\times10^{-5}$ & $5\times10^{-5}$ \\
Language input & None & Task instruction & Task instruction \\
Auxiliary language supervision & None & None & Subtask annotations \\
Auxiliary loss weight & N/A & N/A & $0.1$ \\
Action chunk length & $50$ & $30$ & $30$ \\
Chunk time offsets $k$ & $0$--$49$ & $0$--$29$ & $0$--$29$ \\
Observation history & $1$ & $1$ & $1$ \\
Image resolution & $640\times512$ & $224\times224$ & $224\times224$ \\
\bottomrule
\end{tabular}
}
\end{table}

Image augmentation is applied according to camera viewpoint.
Non-wrist cameras use a random 95\% crop followed by resizing back to the original image size, random rotation within $\pm5^\circ$, and color jitter.
Wrist cameras use color jitter only, with no crop or rotation, to preserve local gripper-object geometry in the close-range manipulation views.

\subsection{\texorpdfstring{$\pi_{0.5}^{+}$}{pi-0.5+} Auxiliary Head}
\label{app:pi_plus}

The $\pi_{0.5}^{+}$ variant augments the default vision-language-action (VLA) backbone with an autoregressive language head trained to predict the current subtask text. At timestep $t$, the target is the verified label $\ell_j$ for which $t_j^{start}\leq t<t_j^{end}$, serialized as \texttt{Subtask: <label>;}. Boundary timestamps are not prediction targets.
The training objective is
\begin{equation}
    \mathcal{L} = \mathcal{L}_{action} + \lambda_{lang}\mathcal{L}_{subtask},
\end{equation}
where $\lambda_{lang}=0.1$ in the main experiments. $\mathcal{L}_{subtask}$ is autoregressive token cross-entropy over the subtask segment; task-prompt, \texttt{Action:}, and padding tokens are excluded from this loss. At inference, action chunks are generated directly without autoregressive subtask decoding.

\section{Evaluation Protocol}
\label{app:evaluation}

\subsection{Task Definitions and Rollout Protocol}
\label{app:task_details}

Table~\ref{tab:app_task_details} summarizes the task randomization and rollout criteria.
Detailed task descriptions are provided in Sec.~\ref{subsec:task_setup}.

\begin{table}[ht]
\centering
\footnotesize
\setlength{\tabcolsep}{3.5pt}
\caption{Task-level evaluation protocol.}
\label{tab:app_task_details}
\resizebox{\linewidth}{!}{
\begin{tabular}{lllc}
\toprule
\textbf{Task} & \textbf{Randomization} & \textbf{Success criterion} & \textbf{Time Limit} \\
\midrule
Pick-and-Place & Object and box poses & Target objects placed in box & 50 s \\
Blocks Collection & Block pose/shape; receptacle pose & Required blocks collected & 180 s \\
Crate Stacking & Crate and target poses & Crate stably stacked & 100 s \\
\bottomrule
\end{tabular}
}
\end{table}

\begin{figure}[H]
    \centering
    \includegraphics[width=\linewidth]{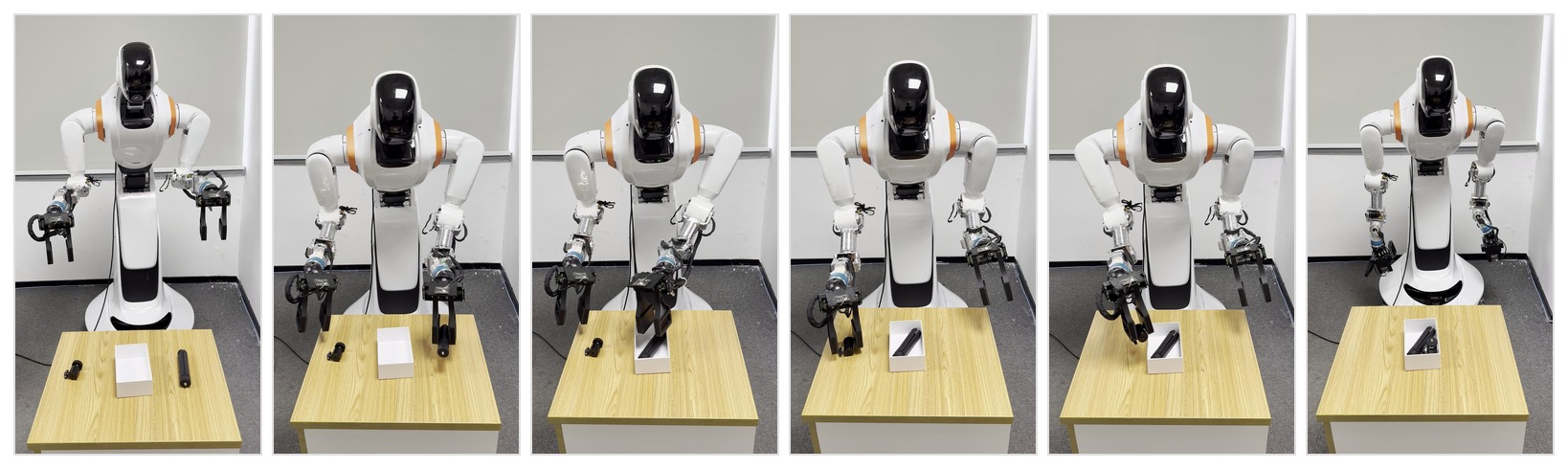}
    \caption{\textbf{Model inference rollout for Task 1.}
    The sequence shows the inferred Pick-and-Place behavior, where the robot approaches the tabletop scene, grasps the objects, and places them into the box.}
    \label{fig:app_task1_rollout}
\end{figure}

\begin{figure}[H]
    \centering
    \includegraphics[width=\linewidth]{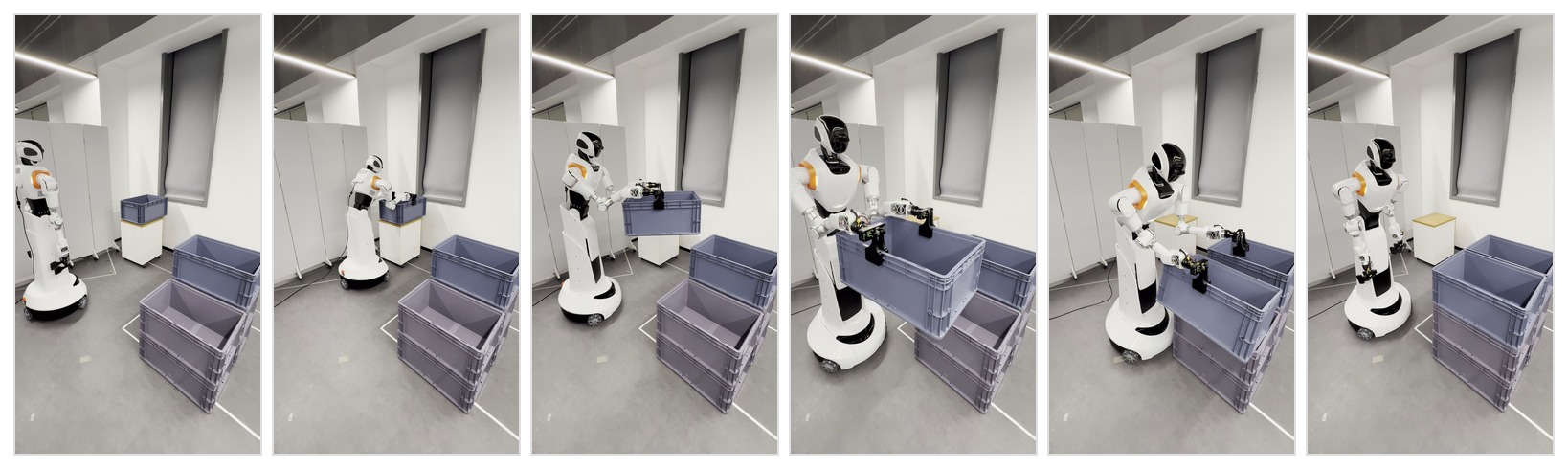}
    \caption{\textbf{Model inference rollout for Task 3.}
    The sequence shows the inferred Crate Stacking behavior, where the robot approaches the source crate, lifts and transports it to the target stack, places it, and retracts after completion.}
    \label{fig:app_task3_rollout}
\end{figure}

Each trained policy is evaluated for 20 trials per task, as reported in the main paper.
Before each trial, the workspace is reset according to the task randomization protocol.
A trial is counted as successful only if the full task is completed within the task-specific time limit.
A trial is counted as failed if the robot drops a required object, reaches an unrecoverable state, violates a safety constraint, or exceeds the time limit.
Figs.~\ref{fig:app_task1_rollout} and~\ref{fig:app_task3_rollout} show example rollouts for Pick-and-Place and Crate Stacking.

\subsection{Throughput and Effective Throughput}
\label{app:throughput_formula}

Collection throughput is reported as valid episodes per measured collection hour:
\begin{equation}
    R_m = \frac{3600N_m}{\sum_{i=1}^{N_m} T_i^{(m)}},
\end{equation}
where $m$ denotes the collection modality and $T_i^{(m)}$ is the measured wall-clock time associated with an episode. For each modality-task cell in Table~\ref{tab:throughput_transposed}, 30 valid demonstrations were collected. The measured interval includes task resets, failed or aborted attempts, robot recovery, synchronous narration, and logging. It excludes one-time tracker-to-torso calibration and joint zeroing (about 10 minutes per hardware configuration), Lighthouse/operator-height setup (about 3 minutes when the workspace or operator changes), fitting and initialization (about 20 seconds per session), and offline transcription, segmentation, and human correction. One expert operator with approximately 1,000 hours of experience collected all conditions. Thus the rate is a steady-state collection measure under the tested setup; it is not an end-to-end rate including all setup and annotation labor.

The effective throughput reported in Table~\ref{tab:effective_throughput} is an autonomous-evaluation metric rather than a data-collection metric. For a policy evaluated over $N$ trials on a task, it is computed as
\begin{equation}
    R_{\text{eff}} =
    \frac{3600\sum_{j=1}^{N}\mathbb{I}[\text{success}_j]}{\sum_{j=1}^{N}\tau_j},
\end{equation}
where $\tau_j$ is the wall-clock evaluation-cycle time for trial $j$, including policy execution, timeout, safety stop, and recovery/reset overhead measured in the evaluation protocol. Equivalently,
\begin{equation}
    R_{\text{eff}} = p\frac{3600}{\bar{\tau}},
\end{equation}
where $p$ is the measured success rate and $\bar{\tau}$ is the mean evaluation-cycle duration. This definition matches the main paper's interpretation of effective throughput as expected successful autonomous completions per hour.

\section{Additional Analyses}
\label{app:additional_analysis}

\subsection{Statistical Reporting}
\label{app:statistics}

Each policy configuration uses one training seed.
Three fixed checkpoints were evaluated separately: 8k, 9k, and 10k steps for both $\pi_{0.5}$ variants, and 40k, 45k, and 50k steps for ACT, each with 20 real-robot rollouts per task.
The main paper reports the best-performing checkpoint for each condition, so each displayed $k/20$ rate has a 20-rollout denominator; the checkpoints are not pooled into $n=60$.
Best-of-three selection biases the reported rates upward, and different conditions may be reported at different checkpoints.
Wilson intervals quantify rollout uncertainty at the reported checkpoint only and do not include seed variation or checkpoint-selection uncertainty.
Table~\ref{tab:app_principal_ci} lists the principal conditions.

\begin{table}[ht]
\centering
\footnotesize
\setlength{\tabcolsep}{3pt}
\caption{Best-checkpoint outcomes (successes/20) and 95\% Wilson intervals.}
\label{tab:app_principal_ci}
\begin{tabular}{lccc}
\toprule
Condition & Pick-Place & Blocks & Stacking \\
\midrule
Teleop 0 / Active 110 & 7 [18.1,56.7] & 2 [2.8,30.1] & 1 [0.9,23.6] \\
Teleop 10 / Active 100 & 17 [64.0,94.8] & 12 [38.7,78.1] & 8 [21.9,61.3] \\
Teleop 110 / Active 0 & 17 [64.0,94.8] & 13 [43.3,81.9] & 8 [21.9,61.3] \\
BRS-style & 16 [58.4,91.9] & 13 [43.3,81.9] & 7 [18.1,56.7] \\
\bottomrule
\end{tabular}
\end{table}

Policies are evaluated on the same trial scenes, so checkpoint-matched comparisons use exact McNemar tests on within-scene discordant outcomes.
The $\Delta q$ plus deadband configuration has 8/8/7 successes at the three checkpoints versus 1/0/0 for absolute joint targets; Holm-adjusted exact tests give $p\leq0.0391$ at each checkpoint.
Deadband-on outcomes (8/8/7) exceed deadband-off outcomes (4/6/4) at every checkpoint, but that comparison is not statistically resolved.
The one-rollout differences between BRS-style and SuperSuit teleoperation are not interpreted as equivalence or superiority.

\subsection{Language Annotation Audit}
\label{app:language_audit}

The audit uses 100 Crate Stacking demonstrations with five human-verified subtasks per episode.
Excluding episode start and end gives four internal transitions per episode, or 400 boundaries.
A proposal is timing-correct when it lies within $\pm1$ second of the final human-corrected timestamp: 327/400 (81.8\%, 95\% Wilson interval [77.7\%,85.2\%]).
This metric does not count missed or extra proposals and is not a complete segmentation precision/recall measure.
Human correction averages 15 seconds per episode with proposals, compared with 37 seconds for fully manual annotation.

\subsection{Cross-Modal Compatibility}
\label{app:cross_modal}

Teleoperation and active demonstrations differ in whether the physical robot executes the task during collection, but both produce the same kinematic target variables.
The shared schema lets one policy action space consume both sources without a modality-specific head or post-hoc trajectory retargeting.
The active-only condition is weaker, so matching action schemas does not by itself close the gap between the two sources.

\subsection{Deadband Failure Mode}
\label{app:deadband_failure}

Without the velocity deadband, low-amplitude operator sway is recorded as base motion.
During autonomous execution, the policy may reproduce this motion even when the grippers are in contact with the object.
In bimanual phases, such drift can change the relative pose between the grippers and the object, break force closure, or accumulate enough chassis displacement to miss the placement target.
This is a plausible mechanism for the lower Crate Stacking point estimates without the deadband (best checkpoint 6/20 versus 8/20 under Teleop 10 / Active 100; 4/6/4 versus 8/8/7 across checkpoints), although that difference is not statistically resolved.

\subsection{Absolute-Joint Failure Mode}
\label{app:absolute_failure}

Absolute joint targets retain any static joint-coordinate offset in the supervised target.
The forward difference cancels such an offset only when the command and the observed state share it, and does not remove time-varying compliance or backlash.
In the Crate Stacking ablation, $\Delta q$ plus deadband has more successes than absolute targets at each of the three checkpoints.

\section{Failure Analysis and Practical Considerations}
\label{app:failure_analysis}

\paragraph{Active-data visual domain gap.}
Active demonstrations show wearable rather than robot arms in the image stream, which is the one difference a reader can check in Fig.~\ref{fig:bimodal_data}.
The remaining differences and their consequences are stated in the limitations.
Arm masking, appearance randomization, or feature alignment could address the visual side in future work.

\paragraph{Tracker visibility and reflective environments.}
The Lighthouse-based tracker depends on stable line-of-sight visibility.
Occlusions, reflective surfaces, and poor base-station placement can introduce pose jumps or short dropouts, which are amplified by velocity differentiation and can appear as unintended base commands.
In practice, base stations should be placed to cover the full operator workspace, reflective clutter should be removed when possible, and tracker confidence or dropout flags should be logged with the demonstration.
Segments with tracking loss or discontinuous pose estimates should be rejected during data filtering.

\paragraph{Embodiment and end-effector scope.}
The wearable was adapted to a second platform with a seven-joint serial arm but different wrist order, frames, and link/end-effector geometry.
Mechanical reconfiguration required approximately 10 person-hours, excluding unattended printing, and preserved direct joint correspondence.
The second embodiment was not included in the policy benchmark, and more substantial kinematic differences would need retargeting.

\end{document}